\documentclass{rAMF2e}
    \usepackage{epstopdf}
    \usepackage{subfigure}
    \usepackage{tikz}
    \usetikzlibrary{positioning}
    \usepackage{caption}
    \usepackage{graphicx}
    \usepackage{float}
    \usepackage{listings}
    \usepackage{algorithm,algpseudocode}
    \usepackage{hyperref}
    \begin{document}

    	\title{Predicting the direction of stock market prices using random forest}
    	\author{
              \begin{tabular}{ccc} 
     Luckyson Khaidem & Snehanshu Saha & Sudeepa Roy Dey\\ 
     khaidem90@gmail.com & snehanshusaha@pes.edu & sudeepar@pes.edu \\
    \end{tabular}
   }

    \maketitle
    \abstract{Predicting trends in stock market prices has been an area of 
            interest for researchers for many years due to its
    		    complex and dynamic nature. Intrinsic volatility in stock 
            market across the globe makes the task of prediction 
            challenging. Forecasting and diffusion modeling, although 
            effective can't be the panacea to the diverse range of problems 
            encountered in prediction, short-term or otherwise. Market 
            risk, strongly correlated with forecasting errors, needs to 
            be minimized to ensure minimal risk in investment. The 
            authors propose to minimize forecasting error by treating  
            the forecasting problem as a classification problem, a 
            popular suite of algorithms in Machine learning. In this 
            paper, we propose a novel way to minimize the risk of 
            investment in stock market by predicting the returns of a 
            stock using a class of powerful machine learning algorithms 
            known as ensemble learning. Some of the technical indicators 
            such as Relative Strength Index 
            (RSI), stochastic oscillator etc are used as inputs to train 
            our model. The learning model used is an ensemble of multiple decision 
            trees. The algorithm is 
            shown to outperform existing algorithms found in the 
            literature. Out of Bag (OOB) error estimates have been found 
            to be encouraging.\\}
        \begin{keywords}
        Random Forest Classifier, stock price 
            forecasting, Exponential smoothing, feature extraction, OOB 
            error and convergence.
        \end{keywords}
        \section{Introduction}
        	Predicting the trends in stock market prices is a very 
    		challenging task due to the many uncertainties involved and
    		many variables that influence the market value in a 
    		particular day such as economic conditions, investors' 
            sentiments towards a particular company, political events 
            etc.
            Because of this, stock markets are susceptible to quick 
            changes,
            causing random fluctuations in the stock price.
            Stock market series are generally dynamic, non-parametric,
            chaotic and noisy in nature and hence, 
            stock market price movement is considered to be a random 
            process with 
            fluctuations which are more pronounced for short time 
            windows. 
            However, some stocks usually tend to develop linear trends 
            over long-term time windows. Due to 
            the chaotic and highly volatile nature of stock behavior, 
            investments in share market comes with high risk. In order 
            to minimize the risk involved, advanced knowledge of stock 
            price movement in the future is required. Traders are more
            likely to buy a stock whose value is expected to increase in 
            the future. On the other hand, traders are likely to refrain 
            from buying a stock whose value is expected to fall in the 
            future. So, there is a need for accurately predicting the 
            trends in stock market prices in order to maximize capital 
            gain and minimize loss.
            Among the major methodologies used to   predict stock price behavior, the 
            following are particularly 
             noteworthy:
         (1) Technical Analysis, (2) 
        Time Series Forecasting (3) Machine Learning and Data 
        Mining (\cite{k}) and (4) modeling and predicting volatility of 
        stocks using differential equations (\cite{diffusion}). This 
        paper mainly focuses on the third approach as the data 
        sets associated with stock market
        prediction problem are too big to be handled with non-data
        mining methods. (\cite{asd})
        \par
        Application of Machine learning models in stock market behavior 
        is quite a 	recent phenomenon. The approach is a departure from 
        traditional forecasting and diffusion type methods. Early models 
        used in stock forecasting involved statistical methods
        such as time series model and multivariate analysis (\cite{stat1}, 
        \cite{stat2}, \cite{stat3}). The stock price 
        movement was treated as a function of time series and solved 
        as a regression problem. However, predicting the exact values of
        the stock price is really difficult due to its chaotic nature and 
        high volatility. Stock prediction performs better when it is 
        treated as classification problem instead of a regression 
        problem. The goal is  to design an intelligent 
        model that learns from 
        the market data using machine learning techniques and forecast 
        the future trends in stock price movement. The predictive output 
        from our model may be used to support decision making for people 
        who invest in stock markets. 
        Researchers have used a variety of algorithms such as SVM, Neural 
        Network, Naive Bayesian Classifier etc. We will discuss the 
        works done by other authors in the next section.
        \section{Related Work}
        The use of prediction algorithms to determine 
        future trends in stock market prices contradict
        a basic rule in finance known as the Efficient Market
        Hypothesis (\cite{EMH}). It states that current stock prices 
        fully reflect all the relevant information. It implies that
        if someone were to gain an 
        advantage by analyzing historical stock 
        data, then the entire market will become aware of this 
        advantage and as a result, the price of the share will be 
        corrected. This is a highly controversial and often 
        disputed theory. Although it is generally accepted, there are
        many researchers who have rejected this theory by using 
        algorithms that can model more complex dynamics of the financial 
        system (\cite{emhcrit}). 
        
        \par 
        Several algorithms have been used in stock prediction 
        such as SVM, Neural Network, Linear Discriminant Analysis, 
        Linear Regression, KNN and Naive Bayesian Classifier. 
        Literature survey revealed that SVM has been used most
        of the time in stock prediction research.  
        \cite{stanford1} have considered sensitivity of stock prices to 
        external condition. The external 
        conditions taken into consideration include daily quotes of
        commodity prices such as gold, crude oil, nature gas,
        corn and cotton in 2 foreign currencies (EUR, JPY). In addition
        to that, they collected daily trading data of 2666 U.S stocks
        trading (or once traded) at NYSE or NASDAQ from 2000-01-01 to
        2014-11-10. This dataset includes everyday open price, 
        close price, highest price, lowest price and trading volume of 
        every 
        stock. Features were derived using the information from the 
        historical stock data as well as external variables which were 
        mentioned earlier in this section. It was 
        found that logistic regression turned out to be the best model with
        a success rate of 55.65\%. In \cite{stanford2}, the training
        data used in their research was 3M Stock data. The 
        data contains daily stock information ranging from 1/9/2008 to 
        11/8/2013 (1471 data points). Multiple algorithms were chosen
        to train the prediction system. These algorithms are Logistic
        Regression, Quadratic Discriminant Analysis, and SVM.
        These algorithms were applied to next day model which 
        predicted the outcome of the stock price on the next day and long 
        term model, which predicted the outcome of the stock price
        for the next $n$ days.
        The next day prediction model produced accuracy results 
        ranging from 44.52\% to 58.2\%. \cite{stanford2} have justified 
        their results by stating that US stock market is semi-strong 
        efficient, meaning that neither fundamental nor technical 
        analysis can be used to achieve superior gain. However, the 
        long-term prediction model produced better results which peaked
        when the time window was 44. SVM reported the highest accuracy of 
        79.3\%. In \cite{stanford3}, the 
        authors have used 3 stocks (AAPL, MSFT, AMZN) that have time 
        span available from 2010-01-04 to 2014-12-10. Various technical indicators 
        such as RSI, On balance 
        Volume, Williams \%R etc are used as features. Out of 84 
        features, an extremely randomized tree algorithm was 
        implemented as described in \cite{randomizedtree}, for the 
        selection of the most relevant features.
        These features were then fed to an rbf Kernelized SVM for 
        training. 
       \cite{devi} has proposed a model which uses hybrid cuckoo search 
       with
       support vector machine (With Gaussian kernel). Cuckoo search 
       method is 
       an optimization 
       technique used to optimize the parameters of support vector 
       machine. 
       The proposed model used technical indicators such as RSI, Money 
       Flow Index, EMA, Stochastic Oscillator and MACD . The data used in 
       the proposed system consists of daily 
       closing prices of BSE-Sensex and CNX - Nifty from Yahoo finance 
       from 
       January 2013 to July 2014. \cite{ANN} proposes a trading agent 
       based
       on a neural network ensemble, that predicts if one stock is going 
       to 
       rise or fall. They evaluated their model in two databases: The 
       North 
       American and the Brazilian stock market. \cite{dss} implemented a 
       One vs 
       All and One vs One neural network to classify Buy, hold or Sell 
       data
       and compared their performance with a traditional neural network. 
       Historical data of Stock Exchange of Thailand (SET)
        of seven years (date 03/01/2007 to 29/08/2014) was
    	selected. It was found that OAA-NN performed better than OAO-NN 
        and 
        traditional NN models, producing an average accuracy of 72.50\%.
       \par
        The literature survey helps us conclude that
        Ensemble learning algorithms have remained unexploited in the
        problem of stock market prediction. We will be using an ensemble 
        learning method known as 
        Random Forest to build our predictive model. Random forest is 
        a multitude of decision trees whose output is the mode of the 
        outputs from the individual trees.
        
      \par  The remainder of the paper is organized as follows. Section $3$ discusses about 
      data and the operations implemented on data that include cleaning, 
      pre-processing, feature extraction, testing for linear separability and learning the 
      data via random 
      forest ensemble. Section $4$ traces the algorithm by using graph 
      description language 
      and computes the OOB error. Section $5$ contains a brief outline on 
      OOB error and 
      convergence estimate. The next section documents the results 
      obtained, followed by a 
      comparative study establishing the superiority of the proposed 
      algorithm. We conclude 
      by summarizing our work in section $7.1$.
      \section{Methodology and Analysis}
         \begin{figure}[H]
    	\begin{center}
    	\captionsetup{justification = centering}
    	\begin{tikzpicture}[mystyle/.style=
        {draw,rectangle,fill=blue!30,thick,minimum
    	width=3cm,minimum height=1cm}]
    	\node[mystyle] (A) {Data Collection};
    	\node[mystyle] (B) [below=of A] {Exponential Smoothing};
    	\node[mystyle] (C) [below=of B] {Feature Extraction}; 
    	\node[mystyle] (D) [below=of C] {Ensemble Learning};
    	\node[mystyle] (E) [below=of D] {Stock Market Prediction};
    	\draw[->] (A) -- (B);
    	\draw[->] (B) -- (C);
    	\draw[->] (C) -- (D);
    	\draw[->] (D) -- (E);
    	\end{tikzpicture}
    	
    	\end{center}
    	\end{figure}
        \begin{center}
        \mbox{Fig 1: Proposed Methodology}
        \end{center}
        The learning algorithm used in our paper is random 
        forest. The time series data is acquired, smoothed and technical 
        indicators are extracted. Technical indicators are 
        parameters which provide insights to the expected stock price 
        behavior in future. These technical indicators are then used
        to train the random forest. The details of each step will be 
        discussed in this section.
    	\subsection{Data Preprocessing}
       The time series historical stock data is first exponentially 
        smoothed. 
        Exponential 
        smoothing applies more weightage to the recent observation and 
        exponentially decreasing 
        weights to past observations.
    	The exponentially smoothed statistic of a series $Y$ can be 
        recursively calculated as:
    	\begin{equation}
    		S_{0} = Y_{0}
    	\end{equation}
     	\begin{equation}
     		\mbox{for $t$ $>$ 0, } S_{t} = \alpha * Y_{t} + (1 - \alpha) * S_{t-1}
    	\end{equation}
        where $\alpha$ is the smoothing factor and 0 $<$ $\alpha$ 
    	$<$ 1. Larger values of $\alpha$ reduce the level of 
    	smoothing. When $\alpha$ = 1, the smoothed statistic becomes
    	equal to the actual observation. The smoothed statistic 
    	$S_{t}$ can be calculated as soon as two observations are 
    	available. This smoothing removes random variation or noise from 
        the historical data allowing the model to easily identify long 	 
        term price trend in the stock price behavior. Technical 
        indicators are then calculated from the exponentially smoothed 
        time series data which are later organized into feature matrix.
        The 
        target to be predicted in the
    	$i^{th}$ day is calculated as follows:
    	\begin{equation}
    		target_{i} = Sign(close_{i+d} - close_{i})
    	\end{equation}
    	where $d$ is the number of days after which the prediction is to 
        be made. When the value of $target_{i}$ is +1,
    	it indicates that there is a positive shift in the price after 
        $d$ days and -1 indicates that there is a negative shift after 
        $d$ days. 
        The $target_{i}$ values are assigned as labels to the $i^{th}$ 
        row in 
        the feature matrix.
        \subsection{Feature Extraction}
        Technical Indicators are important parameters that are 
    	calculated from time series stock data that aim to forecast 
        financial market direction. They are tools which are widely used 
        by investors to check for bearish or bullish signals.
    	The technical indicators which we have used are 
    	listed below\\
    \begin{description}
      \item[Relative Strength Index] \hfill \\ \\
      The formula for caculating RSI is:
      \begin{equation}
        RSI = 100 - \dfrac{100}{1+RS}     
      \end{equation}
      \begin{equation}
          RS = \dfrac{\mbox{\textit{Average Gain Over past 14 days}}}
          {\textit{Average Loss Over past 14 days}}
      \end{equation}
      RSI is a popular momentum indicator which determines
      whether the stock is overbought or oversold. A stock is
      said to be overbought when the demand unjustifiably
      pushes the price upwards. This condition is generallyt
      interpreted as a sign that the stock is overvalued and 
      the price is likely to go down. A stock is said to be 
      oversold when the price goes down sharply to a level 
     below its true value. This is a result caused due
     to panic selling. RSI ranges from 0 to 100 and 
     generally, when RSI is above 70, it may indicate that
     the stock is overbought and when RSI is below 30, it may
     indicate the stock is oversold.\\
      \item[Stochastic Oscillator] \hfill \\ \\
        The formula for calculating Stochastic Oscillator is:
        \begin{equation}
            \%K = 100*\dfrac{(C-L14)}{(H14-L14)} 
        \end{equation}
        where,
         \begin{itemize}
             \item[] C = Current Closing Price
             \item[] L14 = Lowest Low over the past 14 days
             \item[] H14 = Highest High over the past 14 days
         \end{itemize}
         Stochastic Oscillator follows the speed or the
         momentum of the price. As a rule, momentum changes
         before the price changes. It measures the level 
         of the closing price relative to low-high range
         over a period of time.\\
      \item[Williams \%R] \hfill \\ \\
      Williams \%R is calculated as follows:
      \begin{equation}
         \%R = \dfrac{(H14 - C)}{(H14 - L14)}*-100
      \end{equation}
        where,
         \begin{itemize}
             \item[] C = Current Closing Price
             \item[] L14 = Lowest Low over the past 14 days
             \item[] H14 = Highest High over the past 14 days
         \end{itemize}
       Williams \%R ranges from -100 to 0. When its value is
       above -20, it indicates a sell signal and when its
       value is below -80, it indicates a buy signal.\\ 
     \item[Moving Average Convergence Divergence] \hfill \\ \\
     The formula for calculating MACD is:
     \begin{equation}
         MACD = EMA_{12}(C) - EMA_{26}(C)
    \end{equation}
    \begin{equation}
         SignalLine = EMA_{9}(MACD) 
         \end{equation}
     where,
     \begin{itemize}
     \item[] $MACD$ = Moving Average Convergence Divergence
     \item[] $C$ = Closing Price series
     \item[] $EMA_{n}$ = n day Exponential Moving Average 
     \end{itemize}
     EMA stands for Exponential Moving Average. When the MACD
     goes below the SingalLine, it indicates a sell signal. 
     When it goes above the SignalLine, it indicates a buy 
     signal.\\
     \item[Price Rate of Change] \hfill \\ \\
    It is calculated as follows:
        \begin{equation}
           PROC(t) = \dfrac{C(t) - C(t-n)}{C(t-n)}
        \end{equation}
        where,
        \begin{itemize}
        \item[] PROC(t) = Price Rate of Change at time t
        \item[] C(t) = Closing price at time t
        \end{itemize}
        It measures the most recent change in price with 
        respect to the price in n days ago.\\
     \item[On Balance Volume] \hfill \\ \\
     This technical indicator is used to find buying and 
     selling trends of a stock. The formula for calculating
     On balance volume is:
    \begin{equation}
     OBV(t) =
      \begin{cases} 
          \hfill OBV(t-1) + Vol(t)   \hfill & \text{ if $C(t)$ $>$ $C(t-
          1)$} \\
          \hfill OBV(t-1) - Vol(t) \hfill & \text{ if $C(t)$ $<$ $C(t-
          1)$} \\
          \hfill OBV(t-1) \hfill& \text{if $C(t)$ = $C(t-1)$} \\
      \end{cases}
    \end{equation}
    where,
    \begin{itemize}
    \item[] OBV(t) = On Balance Volume at time t
    \item[] Vol(t) = Trading Volume at time t
    \item[] C(t) = Closing price at time t
    \end{itemize}
     \end{description}
    \subsection{Test for linear separability}

       \begin{figure}[H]
        \begin{center}
    \includegraphics[width= 12cm, height = 10cm]{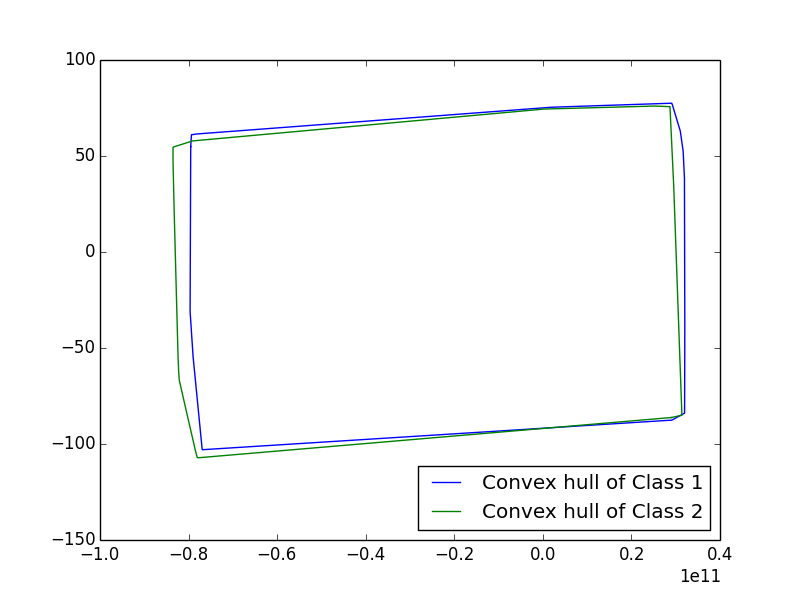}
    \end{center}
    \end{figure}
    \begin{center}
    \mbox{Fig 2: Test for linear separability}
    \end{center}
    Before feeding the training data to the Random Forest 
    	Classifier, the two classes of data are tested for 
    	linear separability by finding their convex hulls. 
    	Linear Separability is
    	a property of two sets of data points where the two sets 
    	are said to be linearly separable if there exists a 
    	hyperplane such that all the points in one set lies 
    	on one side of the hyperplane and all the points in other set 
        lies 
        on the other side of the hyperplane.\\
    	Mathematically, two sets of points $X_{0}$ and 
    	$X_{1}$ in $n$ dimensional Euclidean space
    	are said to be linearly separable if there exists an 
    	$n$ dimensional normal vector $W$
    	of a hyperplane and a scalar $k$, such that every 
    	point $x \in X_{0}$ gives $W^{T}x > k$ 
    	and every point $x \in X_{1}$ gives $W^{T}x < k$. Two
    	sets can be checked for linearly
    	separability by constructing their convex hulls.
    	\par
    	The convex hull of a set of points $X$ is its subset 
    	which forms the smallest convex polygon that
    	contains all the points in $X$. A polygon is said to 
    	be convex if a line joining any two points on the polygon also lies 
        on the polygon. In order to check for Linear Separability, the 
        convex hulls for the two classes are constructed. If the 
    	convex hulls intersect each other, then the classes 
    	are said to be linearly inseparable. 
        Principle component analysis is 
        performed to reduce the 
    	dimensionality of the extracted features into two dimensions. This is
    	done so that the convex hull can be easily visualized in 2 dimensions. 
    	The convex hull test reveals that the classes are not
    	linearly separable as the convex hulls almost overlap. This 
    	observation concludes that Linear Discriminant Analysis
    	cannot be applied to classify our data and hence, providing a
    	stronger justification to why Random Forest Classifier is 
    	used. Another important reason is that since each 
    	decision trees in the forest operate on the random subspace 
    	of the feature space, it leads to the automatic selection of the most 
    	relevant subset of features. Before discussing the RF algorithm, we will be 
        looking at some key definitions in the following section.
        \subsection{Key Definitions}  
        
        Assume there are n data points D = $\{(x_i, y_i)\}^n_{i = 1}$ and feature 
        vectors 
        $\{x_i\}^n_{i = 1}$ with stated outcomes. Each feature vector is d-
        dimensional.\\
       \\
       \textbf{Definition 1: } We define a classification tree where each 
       node is endowed 
       with a binary decision \textit{if $x_i$ $<=$ k  or not ; where $k$ is some 
       threshold}. The topmost node 
       in the classification tree contains all the data points and the set 
       of data is 
       subdivided among the children of each node as defined by the classification
       . The process
       of subdivision continues until every node below has data belonging to one class only.
       Each node is characterized by the feature, $x_i$ and threshold $k$ chosen in such a way
       that minimizes diversity among the children nodes. This is often referred as gini 
       impurity.
       \\ \\
       \textbf{Definition 2: } $X = (X_1,..., X_d)$ is an array of random variables defined 
       on probability space called as random vectors. The joint distribution of $X_1,..., X_d$
       is a measure on $\mu$ on $R^d$, $\mu(A) = P(X \in A)$, $A \in R^d$ 
       where $d = 1,...., 
       m$. For example, Let $x = (x_i,...., x_d)$ be an array of data 
       points. Each feature 
       $x_{i}$ is defined as a random variable with some distribution. Then 
       the random vector 
       $X$ has joint distribution identical to the data points, $x$.
       \\ \\
       \textbf{Definition 3: } Let us represent $h_k(x) =  h(x|\theta_k)$ 
       implying decision tree
     	$k$ leading to a classifier $h_{k}(x)$. Thus, a random forest is a 
        classifier based on 
        a family of classifiers $h(x|\theta_1),...., h(x|\theta_k)$, built 
        on a classification
        tree with model parameters $\theta_k$ randomly chosen from model 
        random vector 
        $\theta$.
        Each classifier, $h_{k}(x) = h(x|\theta_k)$ is a predictor of the 
        number of training 
        samples. $y =^+_-1$ is the outcome associated with input data, $x$ 
        for the final classification function, $f(x)$. 
        \par
        Next, we describe the working of the Random Forest learner by 
        exploiting the key concepts defined above.
        \subsection{Random Forest}
    	Decision trees can be used for various machine learning 
    	applications. But trees that are grown really deep to learn 
    	highly irregular patterns tend to overfit the training sets. A 
    	slight noise in the data may cause the tree to grow in a 
    	completely different manner. This is because of the fact that 
    	decision trees have very low bias and high variance. Random 
    	Forest overcomes this problem by training multiple decision 
    	trees on different subspace of the feature space at the cost of
    	slightly increased bias. This means none of the trees in the forest 
        sees the entire training data. The data is recursively split into 
        partitions. At a particular node, the split is done by asking a 
        question on an attribute. The choice for the splitting criterion is 
        based on some impurity measures such as Shannon Entropy or Gini 
        impurity. 
        \par
        Gini impurity is used as the function 
    	to measure the quality of split in each node. Gini impurity at node
    	N is given by
    	\begin{equation}
    		g(N) = \sum_{i \neq j}P(\omega_{i})P(\omega_{j})
    	\end{equation}
    	where $P(\omega_{i})$ is the proportion of the population with 
        class 
        label i. Another function which can be used to judge the quality of 
        split is Shannon Entropy. It measures
    	the disorder in the information content. In Decision trees, Shannon 
        entropy is used
    	to measure the unpredictability in the information contained in a 
        particular node of a 
    	tree (In this context, it measures how mixed the population in a 
        node is). The entropy in a node $N$ can be calculated as follows
    	\begin{equation}
    		H(N) = -\sum_{i = 1}^{i = d}P(\omega_{i})log_{2}(P(\omega_{i}))
    	\end{equation}
    	where $d$ is number of classes considered and $P(\omega_{i})$ is 
        the 
        proportion of
    	the population labeled as $i$. Entropy is the highest when all the 
        classes are contained
    	in equal proportion in the node. It is the lowest when there is 
        only 
        one class present in
    	a node (when the node is pure).
    	\par
    	The obvious heuristic approach to choose the best splitting decision 
        at a node is the one that reduces the impurity as much as possible. 
        In order words, the best split is characterized by the highest gain 
        in information or the highest reduction in impurity.
    	The information gain due to a split can be calculated as follows
    	\begin{equation}
    		\Delta I(N) = I(N) - P_{L}*I(N_{L}) - P_{R}*I(N_{L})
    	\end{equation} 
    	where $I(N)$ is the impurity measure (Gini or Shannon Entropy) of 
        node $N$, $P_{L}$ is
    	the proportion of the population in node $N$ that goes to the left 
        child of $N$
    	after the split and similarly, $P_{R}$ is the proportion of the 
        population in node
    	$N$ that goes to the right child after the split. $N_{L}$ and 
        $N_{R}$ 
        are the left and right child of $N$ respectively.
    	\par
    	At the heart of all ensemble machine learning algorithms is 
        Bootstrap 
        aggregating, also
    	known as bagging. This method improves the stability and accuracy 
        of 
        learning algorithms.
    	At the same time, it also reduces variance and overfitting which is 
        a common problem while
    	constructing Decision trees.Given a sample dataset $D$ of size $n$, 
        bagging generates $B$ new sets of size $n^{'}$
    	by sampling uniformly from $D$ with replacement. With this 
        knowledge, we can now summarize
    	the algorithm of random forest classifier as follows
    	\begin{algorithm}[H]
      \caption{Random Forest Classifier}\label{RF}
      \begin{algorithmic}[1]
        \Procedure{RandomForestClassifier}{$D$}\Comment{D is the labeled 
        training data}
        \State \texttt{$forest$ = new Array()}
          \For \texttt{ $i$ = 0 to B}
            \State \texttt{$D_{i}$ = Bagging($D$)} \Comment{Bootstrap 
            Aggregation}
            \State \texttt{$T_{i}$ = new DecisionTree()} 
            \State \texttt{$features_i$ = RandomFeatureSelection($D_i$)}
            \State \texttt{$T_{i}$.train($D_{i}$,$features_i$)}
            \State \texttt{$forest$.add($T_{i}$)}
        \EndFor
          \State \textbf{return} $forest$
        \EndProcedure
      \end{algorithmic}
    \end{algorithm}
    \section{Tracing the RF algorithm}
    In this section we will trace the Random Forest algorithm for a 
    particular test sample. To begin with, we trained a random forest using 
    the 
    Apple Dataset for a time window of 30 days. We generated graph 
    description
    language files describing the forest. The output of this 
    process is 30 
    .dot files that corresponds to 30 decision trees in the random 
    forest. 
    These files are found in found in \url{https://drive.google.com/open?
    id=0B980lHZhHCf1Y0s1Q3AwbjVCWGM}. Next, we wrote a python script that 
    reads all the .dot files 
    and 
    traces the RF algorithm for a test sample.
    \subsection{Graph Description Language}
    Graph Description Language is a structured language that is used to 
    describe graphs
    that can be understood both by humans and computers. It can be used to 
    describe both
    directed and undirected graphs. A graph description language begins 
    with 
    the \textit{graph}
    keyword to define a new graph and the nodes are defined within curly 
    braces. The relationship
    between nodes are specified using double hyphen (--) for an undirected 
    graph and arrows (->)
    for a directed graph. The following is an example examples of a Graph 
    Description Language. \\
    \begin{lstlisting}
    graph graphname {
         a -- b -- c;
         b -- d;
     }
    \end{lstlisting}
    \begin{figure}[h]
    \begin{center}
    \includegraphics[width=3 cm, height = 4cm]{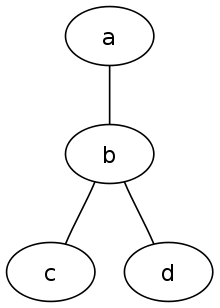}
    \end{center}
    \begin{center}
    \mbox{Fig 3: An undirected graph}
    \end{center}
    \end{figure}
    \subsection{Trace output}
    For the sake of convenience, we will be showing the trace of only 3 
    trees out of 30 trees in the forest. For trace of the entire forest, 
    check: 
    \url{https://drive.google.com/open?id=0B980lHZhHCf1T3dvNDJsVzFfaFE}.
    We take a test sample with the following features and run our trace
    script.
    \begin{itemize}
    \item RSI: 91.638968318801957
    \item Stochastic Oscillator: 88.201032228068314
    \item Williams: -11.798967771931691
    \item Moving Average Convergence Divergence: 5.9734426013145026
    \item Price Rate of Change: 0.11162583681857041
    \item On Balance Volume: 6697423901.3580704
    \end{itemize}
    For Tree 0:\\
    At node 0:(MACD=5.97344260131) \textless= -8.6232?\\
    False: Go to Node 10 \\
    At node 10:(Stochastic Oscillator=88.2010322281) \textless= 80.9531?\\
    False: Go to Node 134 \\
    At node 134:(MACD=5.97344260131) \textless= 10.2566?\\
    True: Go to Node 135 \\
    At node 135:(RSI=91.6389683188) \textless= 97.9258?\\
    True: Go to Node 136 \\
    At node 136:(MACD=5.97344260131) \textless= -1.7542?\\
    False: Go to Node 140 \\
    At node 140:(Price Rate Of Change=0.0412189145804) \textless= 0.0708?\\
    True: Go to Node 141 \\
    At node 141:(MACD=5.97344260131) \textless= 9.2993?\\
    True: Go to Node 142 \\
    At node 142:(MACD=5.97344260131) \textless= 7.7531?\\
    True: Go to Node 143 \\
    At node 143:(On Balance Volume=23722858211.2) \textless= 24938491904.0?\\
    True: Go to Node 144 \\
    At node 144:(Williams=-11.7989677719) \textless= -15.7228?\\
    False: Go to Node 154 \\
    Leaf Node 154 is labeled as Rise \\
    \\
    \\
    For Tree 1:\\
    At node 0:(MACD=5.97344260131) \textless= -8.3677?\\
    False: Go to Node 8 \\
    At node 8:(RSI=91.6389683188) \textless= 42.5156?\\
    False: Go to Node 56 \\
    At node 56:(MACD=5.97344260131) \textless= 9.888?\\
    True: Go to Node 57 \\
    At node 57:(Price Rate Of Change=0.0412189145804) \textless= 0.0612?\\
    True: Go to Node 58 \\
    At node 58:(Stochastic Oscillator=88.2010322281) \textless= 80.3201?\\
    False: Go to Node 108 \\
    At node 108:(On Balance Volume=23722858211.2) \textless= 19212566528.0?\\
    False: Go to Node 110 \\
    At node 110:(Price Rate Of Change=0.0412189145804) \textless= -0.0245?\\
    False: Go to Node 112 \\
    At node 112:(Price Rate Of Change=0.0412189145804) \textless= 0.0554?\\

    True: Go to Node 113 \\
    At node 113:(On Balance Volume=23722858211.2) \textless= 19823968256.0?\\
    False: Go to Node 115 \\
    Leaf Node 115 is labeled as Rise \\
    \\
    \\
    For Tree 2:\\
    At node 0:(Stochastic Oscillator=88.2010322281) \textless= 13.7965?\\
    False: Go to Node 10 \\
    At node 10:(On Balance Volume=23722858211.2) \textless= 26287554560.0?\\
    True: Go to Node 11 \\
    At node 11:(Price Rate Of Change=0.0412189145804) \textless= -0.0649?\\
    False: Go to Node 19 \\
    At node 19:(Stochastic Oscillator=88.2010322281) \textless= 80.6269?\\
    False: Go to Node 121 \\
    At node 121:(MACD=5.97344260131) \textless= 12.8322?\\
    True: Go to Node 122 \\
    At node 122:(MACD=5.97344260131) \textless= 3.3398?\\
    False: Go to Node 132 \\
    At node 132:(Price Rate Of Change=0.0412189145804) \textless= 0.0816?\\
    True: Go to Node 133 \\
    At node 133:(Price Rate Of Change=0.0412189145804) \textless= 0.08?\\
    True: Go to Node 134 \\
    At node 134:(RSI=91.6389683188) \textless= 97.229?\\
    True: Go to Node 135 \\
    Leaf Node 135 is labeled as Rise \\
    \par
    29 of the trees in forest predict a rise in price while a single tree 
    predicts a fall in price. As a result, the output of the ensemble is 
    Rise. This prediction matches with the 
    actual label assigned to the test sample. Each tree 
    recursively divides the feature space into multiple partitions and each 
    partition is 
    given a label that indicates whether the closing price will rise or 
    fall
    after 30 days. Looking at the decision trees in the 
    forest, it is really hard to fathom why the data is split on a particular 
    attribute, 
    especially when the same attribute may be used to split the data 
    further 
    down along the tree. To understand why a particular split is chosen at 
    a 
    node, we need to to be familiar with the concept of impurity measures 
    such 
    as Shannon Entropy and Gini impurity. 
    The decision rules learned by the trees may not be easily understood 
    due 
    to complexities in the underlying pattern of the training data. This 
    is 
    where random forests lose some favor with a technically minded person 
    who 
    likes to know what is under the hood.
    \par 
    It should be noted here that our algorithm converges as the 
    number of trees in the forest increases. We calculated out of bag 
    (OOB) error
    of the classifier with respect to the apple dataset for proof of 
    convergence. In the table given below, the first column indicates the 
    the time window after which the prediction is to be made, the second 
    column indicates the number of trees in the forest, the third column
    indicates the size of the training sample used and the last column 
    is the OOB error rate. 
    \begin{figure}[H]
     \begin{center}
     \begin{tabular}{| c | c | c | c | c |} 
     \hline
     Trading Period (Days) & No. of Trees & Sample Size & OOB error\\
     [0.5ex] 
     \hline\hline
     30 & 5 & 6590 &  0.241729893778 \\ 
     \hline
     30 & 25 & 6590 & 0.149165402124 \\
     \hline
     30 & 45 & 6590 & 0.127617602428  \\
      \hline
     30 & 65 & 6590 & 0.123672230653  \\
      \hline
     60 & 5 & 6545 & 0.198472116119 \\
      \hline
     60 & 25 & 6545 &  0.0890756302521 \\
     \hline
     60 & 45 & 6545 &   0.0786860198625 \\
     \hline
     60 & 65 & 6545 & 0.0707410236822\\
     \hline
     90 & 5 & 6500 & 0.191384615385 \\
     \hline
    90 & 25 & 6500 & 0.0741538461538 \\
    \hline
    90 & 45& 6500 & 0.0647692307692 \\
    \hline
    90 & 65 & 6500 & 0.0555384615385 \\
     \hline
    \end{tabular}
    \end{center}
    \end{figure}
    \begin{center}
    \mbox{Fig 4: OOB error calcuation}
    \end{center}
    As observed, the error rate decreases as the number of trees in 
    the forest is increased. More details about error rates and 
    convergence will be discussed in the next section.
    \section{OOB error and Convergence of the Random Forest}
    Given an ensemble of decision trees $h_{1}(X), h_{2}(x), h_{3}
    (x),...,h_{k}(x)$, as in \cite{rf} we define margin function as
    \begin{equation}
    mg(X,Y) = av_{k}I(h_{k}(X)=Y) - max_{j \neq Y}av_{k}I(h_{k}(X)=j)
    \end{equation}
    where $X,Y$ are randomly distributed vectors from which the training 
    set is drawn. Here, $I(.)$ is the indicator function. The generalization error is given by
    \begin{equation}
    PE^{*} = P_{X,Y}(mg(X,Y) < 0)
    \end{equation}
    The $X$ and $Y$ subscripts indicate that probability is calculated
    over $X,Y$ space.
    In random forests, the kth decision tree $h_{k}(x)$ can be represented
    as $h(x,\theta_{k})$ where $x$ is the input vector and $\theta_{k}$ is 
    the bootstrapped dataset which is used to train the kth tree.
    For a sequence of bootstrapped sample sets $\theta_{1}, \theta_{2}, .
    .., \theta_{k}$ which are generated from the original dataset $\theta$,
    it is found that $PE^{*}$ converges to 
    \begin{equation}
    P_{X,Y}(P_{\theta}(h(X,\theta)=Y) - max_{j\neq Y} P_{\theta}(h(X,Θ)=j)<0)
    \end{equation}
    The proof can be found in appendix I in \cite{rf}.
    To practically prove this theorem with respect to our dataset, the 
    generalization error is estimated using out of bags estimates 
    \cite{oob}. 
    The out of Bag (OOB) error measures the prediction error of Random 
    forests  algorithm and other machine learning algorithms which are 
    based on Bootstrap aggregation.\\
    \textbf{Note:} The average margin of the ensemble of classifiers is the 
    extent to which the average vote count for the correct class flag 
    exceeds the count for the next best class flag.
    \subsection{Random forest as ensembles: An analytical exploration}
    As defined earlier, a Random Forest model specifies $\theta$ as classification tree  marker
    for $h(X|\theta)$ and a fixed probability distribution for $\theta$ for diversity 
    determination in trees is known.\\
    The margin function of an RF is:
    \begin{equation}
    margin_{RF}(x,y) = P_\theta(h(x|\theta) = y) - max_{j \neq y}P_\theta(h(x|\theta) = j)
    \end{equation}
    The strength of the forest is defined as the expected value of the margin:
    \begin{equation}
    s = E_{x,y}(margin_{RF}(x,y))
    \end{equation}
    The generalization error is bounded above by Chebyshev's inequality and is given as
    \begin{equation}
    Error = P_{x,y}(margin_{RF}(x,y) < 0) \leq P_{x,y}(|margin_{RF}(x,y) - s| \geq s) \leq \dfrac{var(margin_{RF}(x,y))}{s^2}
    \end{equation}
    \textbf{Remark: } We know that the average margin of the ensemble of 
    classifiers is 
    the extent to which the average vote count for the correct class flag 
    exceeds the count for the next best class flag. The Strength of 
    the forest is the expected value of this margin. When the margin 
    function gives a negative value, it means that an error has been made
    in classification. The generalization error is the probability that the
    margin is a negative value. Since margin itself is a random variable, 
    equation (20) shows that it is bounded above by its variance divided
    by the square of the threshold. As the strength of the forest grows, error in classification decreases.
    \par
    we present below, the Chebyshev's inequality as the inspiration for the error bound.\\
    \\
    \textbf{Chebyshev's inequality}:
    Let $X$ be any random variable (not necessarily non-negative) and C $>$ 0. Then,
    \begin{equation}
    	P(|X-E(X)| \geq c) \leq \dfrac{var(x)}{c^2}
    \end{equation}
    \textbf{Remark: } It's easy to relate the inequality to the error bound 
    of the Random Forest learner.
    \\ \\
    \textbf{Proof of Chebyshev's inequality: } We require a couple of definitions before the 
    formal proof.
    \begin{itemize}
    \item[] \textbf{A) Indicator Random Variable}
    \begin{equation}
    I(X \geq c) = 
      \begin{cases} 
          \hfill 1   \hfill & \text{ if $X \geq c$} \\
          \hfill 0 \hfill & \text{ Otherwise} \\
         
      \end{cases}
    \end{equation}
    \item[] \textbf{B) Measurable space}
    \begin{equation}
    	A = \{ x \in \Omega | X(x) \geq c\}
    \end{equation}
    \begin{equation}
    	E(X) = \sum_{x \in A}P(x)X(x) = \mu
    \end{equation}
    \end{itemize}
    \textbf{Proof:} 
    \[
    	\mbox{Define}, A = \{x \in \Omega | X(x) - E(x) \geq c\}
     \]
     \[
        \mbox{Thus, } var(X) = \sum_{x \in \Omega}P(X = x)(X(x) - E(x))^2
    \]
    \[
    	= \sum_{x \in A}P(X = x)(X(x) - E(x))^2 + \sum_{x \not\in A}P(X = 
        x)(X(x) - E(x))^2 \geq 0
    \]
    \[
    	\geq \sum_{x \in A}P(x = x)(X(x) - E(x))^2
    \]
    \[
    	\geq P(X=x)c^2 \mbox{\quad since,} X(x) - E(x) \geq C; \forall x \in A
    \]
    \[
    	=c^2 P(A) = c^2 P(|X - E(X)| \geq c)
    \]
    \[
    	=> \dfrac{var(X)}{c^2} \geq P(|X - E(X)| \geq c)
    \]
    \textbf{Remark: } This means that the probability of the deviation of
    a data point from its expected value being greater than $c$, a 
    threshold, is bounded above by the variance of the data points 
    divided
    by the square of the threshold, c. As c increases, the upper bound 
    decreases which implies the probability of a large deviation of a data 
    point from its expected value is less likely. 

    \subsection{OOB error visualization}
    After creating all the decision trees in the forest, for each 
    training
    sample $Z_{i} = (X_{i},Y_{i})$ in the original training set $T$, we 
    select all bagged sets $T_{k}$ which does not contain $Z_{i}$. This 
    set contains bootstrap datasets which do not contain a particular
    training sample from the original training dataset. These sets are 
    called out of bags examples. There are $n$ such sets for each $n$
    data samples in the original training dataset. OOB error is the 
    average error for each $Z_{i}$ calculated using predictions from the 
    trees that do not contain $z_i$ in their respective bootstrap sample.
    OOB error is an estimate of generalization error which measures how
    accurately the random forest predicts previously unseen data. We 
    plotted the OOB error rate for our random forest classifier using the 
    AAPL dataset. 
    \begin{figure}[h]
    \begin{center}
    \includegraphics[width=8 cm, height = 8cm]{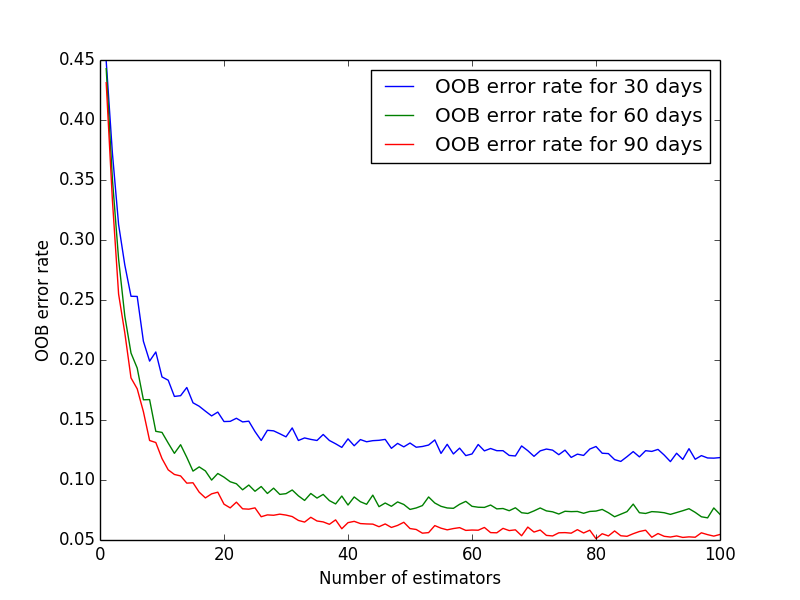}
    \end{center}
    \end{figure}
    \begin{center}
    \mbox{Fig 5: OOB error rate vs Number of estimators}
    \end{center}
    From the above plot, we can see that the OOB error rate decreases
    dramatically as more number of trees are added in the forest. However,
    a limiting value of the OOB error rate is reached eventually.
    The plot shows that the Random Forest converges as more number of
    trees are added in the forest. This result also explains why random 
    forests do not overfit as more number of trees are added into the 
    ensemble.
    \section{Results}
        Using the prediction result produced by our model we can 
        decide whether to buy or sell our stock. If the prediction
        is +1, which means the price is expected to rise after $n$ 
        days then the suggested trading decision is to buy the stock.
        Whereas, if the prediction is -1, it means the price is 
        expected to fall after $n$ days and the suggested trading 
        decision is to sell the stock. Any wrong prediction can
        cause the trader a great deal of money. Hence, the model should
        be evaluated for its robustness. The parameters that are used
        to evaluate the robustness of a binary classifier are 
        accuracy, precision, recall (also known as sensitivity) and 
        specificity. The formula to calculate these parameters are 
        given below:
        \begin{equation}
            Accuracy = \dfrac{tp+tn}{tp+tn+fp+fn}
        \end{equation}
         \begin{equation}
            Precision = \dfrac{tp}{tp+fp}
        \end{equation}
         \begin{equation}
            Recall = \dfrac{tp}{tp+fn}
        \end{equation}
         \begin{equation}
            Specificity = \dfrac{tn}{tn+fp}
        \end{equation}
        where,
        \begin{itemize}
            \item[] tp = Number of true positive values
            \item[] tn = Number of true negative values
            \item[] fp = Number of false positive values
            \item[] fn = Number of false negative values
        \end{itemize}
        Accuracy measures the portion of all testing samples classified
        correctly. Recall (also known as sensitivity) measures the 
        ability
        of a classifier to correctly identify positive labels while
        specificity measures the classifier's ability to correctly 
        identify
        negative labels. And precision measures the  proportion of all 
        correctly identified samples in a population of samples which are
        classified as positive labels.
        We calculate these parameters for the next 1 Month, 2 Months
        and 3 Months prediction model using AAPL, GE dataset (Which are
        listed on NASDAQ) and Samsung Electronics Co. Ltd. (Which is 
        traded
        in Korean Stock Exchange).
        The results are provided in the tables below:
       \begin{figure}[H]
     \begin{center}
     \begin{tabular}{| c | c | c | c | c |} 
     \hline
     Trading Period & Accuracy\% & Precision & Recall & Specificity \\ 
     [0.5ex] 
     \hline\hline
     1 month & 86.8396 & 0.881818 &  0.870736 & 0.865702\\ 
     \hline
     2 months & 90.6433 & 0.910321 & 0.92599 & 0.880899 \\
     \hline
     3 months & 93.9664 & 0.942004 & 0.950355 &  0.926174\\
     \hline
    \end{tabular}

    \end{center}
    \end{figure}
    \begin{center}
    \mbox{Fig 6: Results for Samsung dataset}
    \end{center}
      \begin{figure}[H]
      \begin{center}
     \begin{tabular}{| c | c | c | c | c |} 
     \hline
     Trading Period & Accuracy\% & Precision & Recall & Specificity \\ 
     [0.5ex] 
     \hline\hline
     1 month & 88.264 & 0.89263 & 0.90724 & 0.84848\\ 
     \hline
     2 months & 93.065 & 0.94154 & 0.93858 & 0.91973 \\
     \hline
     3 months & 94.533 & 0.94548 & 0.96120 &  0.92341\\
     \hline
    \end{tabular}

    \end{center}
    \end{figure}
    \begin{center}
    \mbox{Fig 7: Results for Apple Inc. dataset}
    \end{center}

    \begin{figure}[H]
     \begin{center}
     \begin{tabular}{| c | c | c | c | c |} 
     \hline
     Trading Period & Accuracy\% & Precision & Recall & Specificity \\ 
     [0.5ex] 
     \hline\hline
     1 month & 84.717 & 0.85531 &  0.87637 & 0.80968\\ 
     \hline
     2 months & 90.831 & 0.91338 & 0.93099 & 0.87659 \\
     \hline
     3 months & 92.543 & 0.93128 & 0.94557 &  0.89516\\
     \hline
    \end{tabular}
    \end{center}
    \end{figure}
    \begin{center}
    \mbox{Fig 8: Results for GE Dataset}
    \end{center}

    \subsection{Receiver Operating Characteristic}
    The Receiver Operating Characteristic is a graphical method to 
    evaluate the performance of a binary classifier. A curve is drawn by 
    plotting True Positive Rate (sensitivity) against False Positive Rate (1 - 
    specificity) at various threshold values. ROC curve shows the trade-
    off 
    between sensitivity and specificity. When the curve comes 
    closer to the left-hand border and the top border of the ROC
    space, it indicates that the test is accurate. The closer the 
    curve is to the top and left-hand border, the more accurate
    the test is. If the curve is close to the 45 degrees diagonal 
    of the ROC space, it means that the test is not accurate.
    ROC curves can be used to select the optimal model and 
    discard the suboptimal ones.
    \begin{figure}[H] 
    \begin{center}
    \includegraphics[width=8cm, height = 8cm]{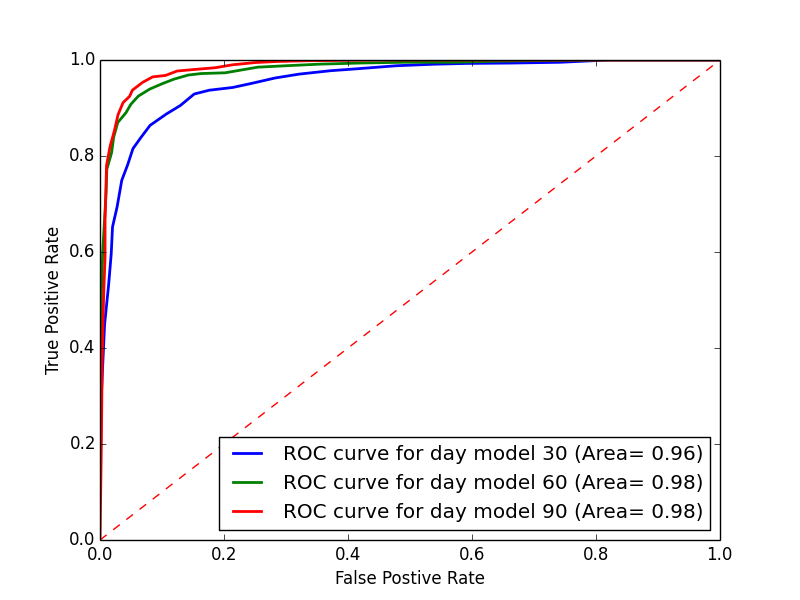}
    \mbox{Fig 9: ROC curves corresponding to AAPL dataset}
    \end{center}
    \end{figure}

    \begin{figure}[H]
    \begin{center}
    \includegraphics[width=8 cm, height = 8cm]{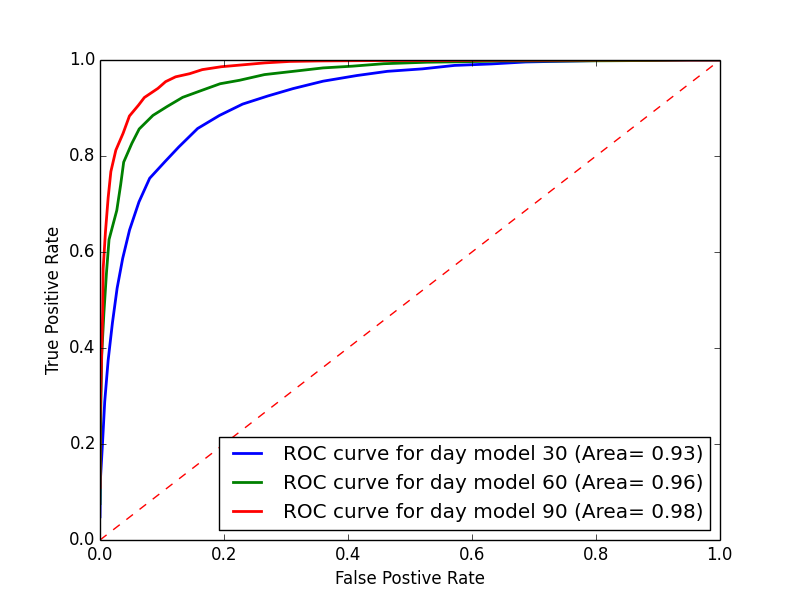}
    \mbox{Fig 10: ROC curves corresponding to GE dataset}
    \end{center}
    \end{figure}

    \begin{figure}[H]
    \begin{center}
    \includegraphics[width=8cm, height = 8cm]{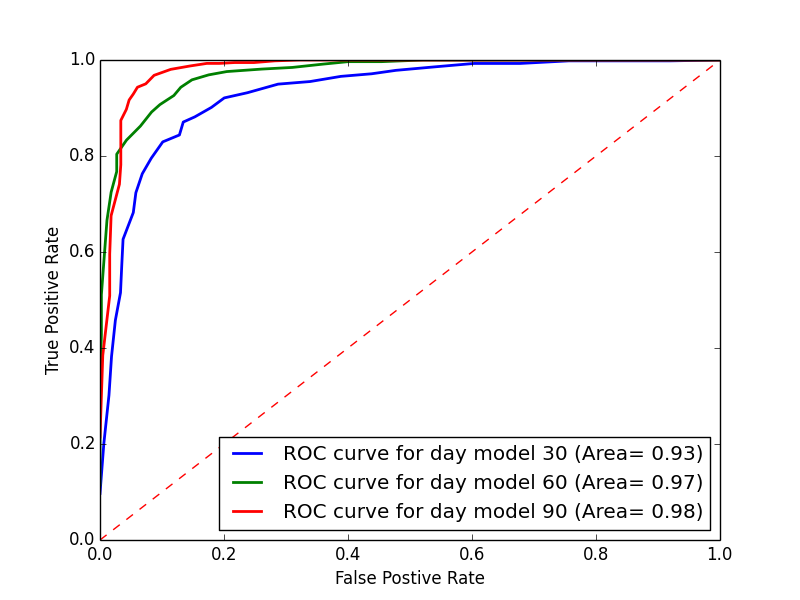}
    \mbox{Fig 11: ROC curves corresponding to Samsung dataset}
    \end{center}
    \end{figure}

    As we can see from the ROC curves, the 90 Day model proves to be the
    most optimal model. The area under the ROC curve is a really important
    parameter for evaluating the performance of a binary classifier. Accuracy 
    is measured by the area under the ROC curve. An area of 1 represents an 
    excellent classifier; an area of .5 represents a worthless classifier 
    which produces random outputs. In other words 
    , the area measures discrimination, that is, the ability of the classifier 
    to correctly classify a positive shift and a negative shift in stock 
    prices in case of our problem. T
    he area under the ROC 
    curve is above 0.9 for all three models using all three datasets. This 
    means our classifier is excellent.

    \section{Discussion and Conclusion}
    The robustness and accuracy of the proposed algorithm in contrast with he 
    ones present in literature need to be discussed. We'll perform a comparative 
    analysis between the results found in \cite{stanford2} and \cite{stanford3} 
    with the 
    results produced by our model for the same dataset. In \cite{stanford2},
    the authors selected 3M stock which contained daily data ranging from
    1/9/2008 to 11/8/2013. They have used four supervised learning 
    algorithms, i.e Logistic Regression, Gaussian Discriminant Analysis, 
    Quadratic Discriminant Analysis, and SVM. Their results are 
    summarized in the Fig 12.
    \par
    \begin{figure}[H]
    \begin{center}
    \includegraphics[width=\linewidth]{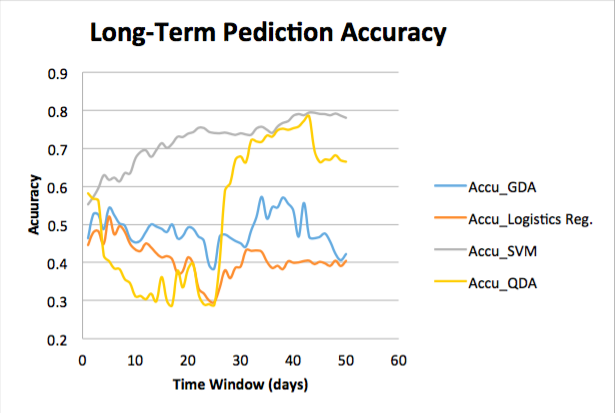}
    \end{center}
    \end{figure} 
    \begin{center}Fig 12: Results from \cite{stanford2} \end{center}
    \par

    \begin{figure}[H]
    \begin{center}
    \includegraphics[width=\linewidth]{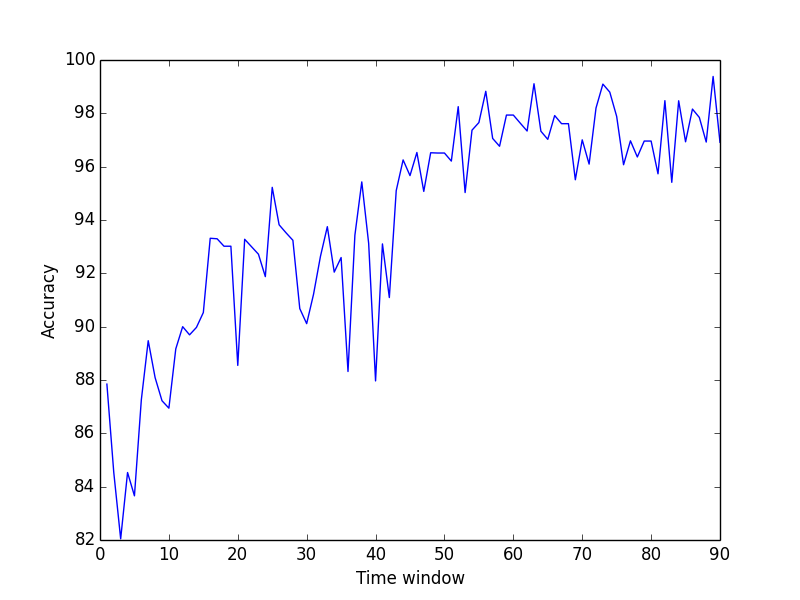}
    \end{center}
    \end{figure}

        \begin{center}Fig 13: Results for 3M stock obtained 
        with our model\end{center}
       
    From the plot in Fig 12:, we can see that for SVM and QDA model, the 
    accuracy increases when the time window increases. Furthermore, SVM 
    gives the highest accuracy when the time window is 44 days (79.3\%). 
    It’s also the most stable model.
    \par 
    Using the same dataset as in \cite{stanford2}, we calculated the accuracy
    for various time widows using the model we have built. The results which 
    we have found is visualized as a graph in Fig 6. As we can see from
    this graph, the accuracy peaked at 96.92\% when the time window is 
    88 days. This is clearly a better result than the one found in 
    \cite{stanford2}. 
    \par 
    For our next comparative, we will be looking at the result obtained in
    \cite{stanford3}. The author of \cite{stanford3} have chosen three
    datasets for the study, i.e AAPL, MSFT and AMZN. The time span of the
    stock data ranges from 2010-01-04 to 2014-12-10. \cite{stanford3} has
    used an extremely randomized tree algorithm to select a subset of
    features from a total of 84 technical indicators. These features
    are then fed to an SVM with rbf kernel for training to predict the next
    3-day, next 5-day, next 7-day and next 10-day trend.
    The results are given in the table below.

    \begin{figure}[H]
    \begin{center}
     \begin{tabular}{| c | c | c | c | c |} 
     \hline
     Company/Accuracy & Next 3-day & Next 5-day & Next 7-day & Next 10-day \\ 
     [0.5ex] 
     \hline
     Apple & 73.4\% & 71.41\% &  70.25\% & 71.13\%\\ 
     \hline
     Amazon & 63\% & 65\% & 61.5\% & 71.25\% \\
     \hline
     Microsoft & 64.5\% & 73\% & 77.125\% &  77.25\%\\
     \hline
    \end{tabular}
    \\
    \mbox{Fig 14: Results from \cite{stanford3}}
    \end{center}
    \end{figure}
    We calculated the accuracy result for the same datasets using our 
    prediction model and obtained the results which given in fig 15.

    \begin{figure}[H]
    \begin{center}
     \begin{tabular}{| c | c | c | c | c |} 
     \hline
     Company/Accuracy & Next 3-day & Next 5-day & Next 7-day & Next 10-day \\ 
     [0.5ex] 
     \hline
     Apple & 85.197\% & 83.88\% &  88.11\% & 92.08\%\\ 
     \hline
     Amazon & 86.51\% & 88.49\% & 85.14\% & 87.46\% \\
     \hline
     Microsoft & 84.59\% & 83.88\% & 89.47\% &  86.46\%\\
     \hline
    \end{tabular}
    \\
    \mbox{Fig 15: Results obtained using our model}
    \end{center}
    \end{figure}
    \cite{devi} used BSE-SENSEX and CNX-NIFTY datasets to predict next day
    outcome using SVM with Cuckoo Search optimization. The results are 
    summarized in the bar charts below.

    \begin{figure}[H] 
    \begin{center}
    \includegraphics[width=5cm, height = 5cm]{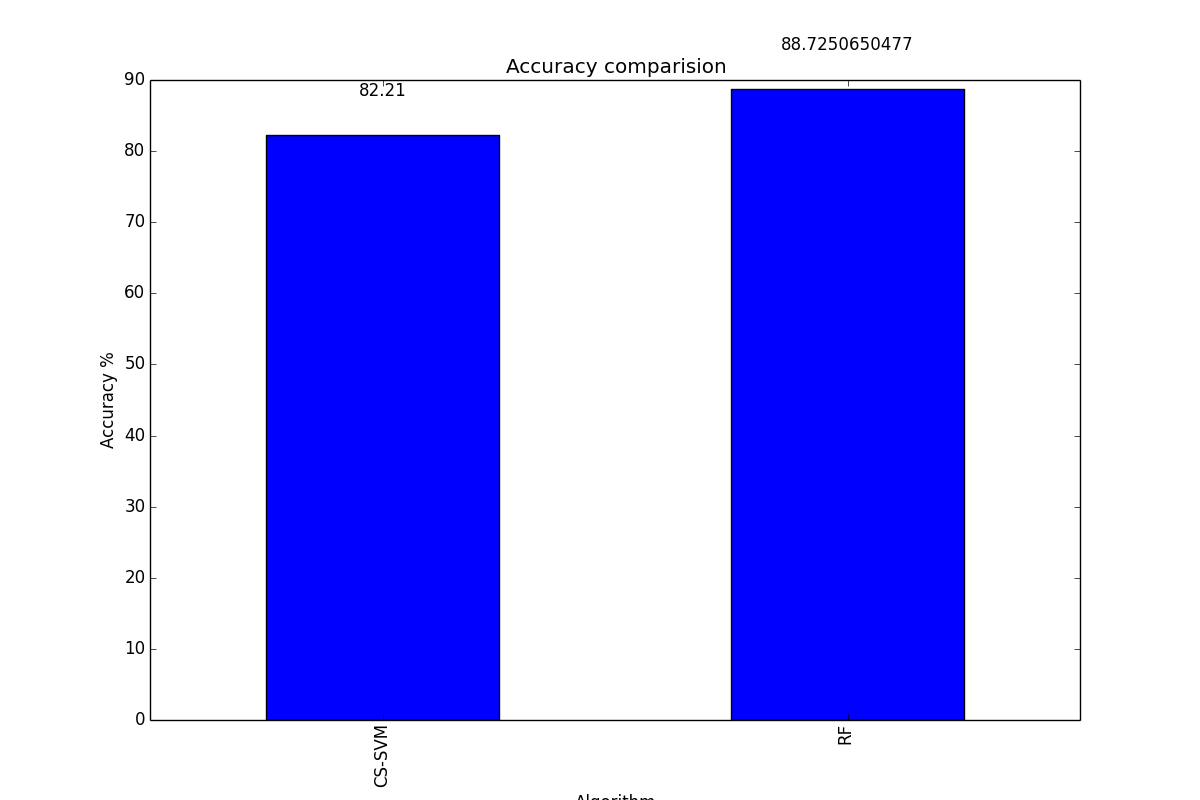}
    \mbox{Fig 16: Comparing Accuracies of CS-SVM as obtained in Devi(2015) and 
    RF for BSE-SENSEX}
    \end{center}
    \end{figure}

    \begin{figure}[H]
    \begin{center}
    \includegraphics[width=5cm, height = 5cm]{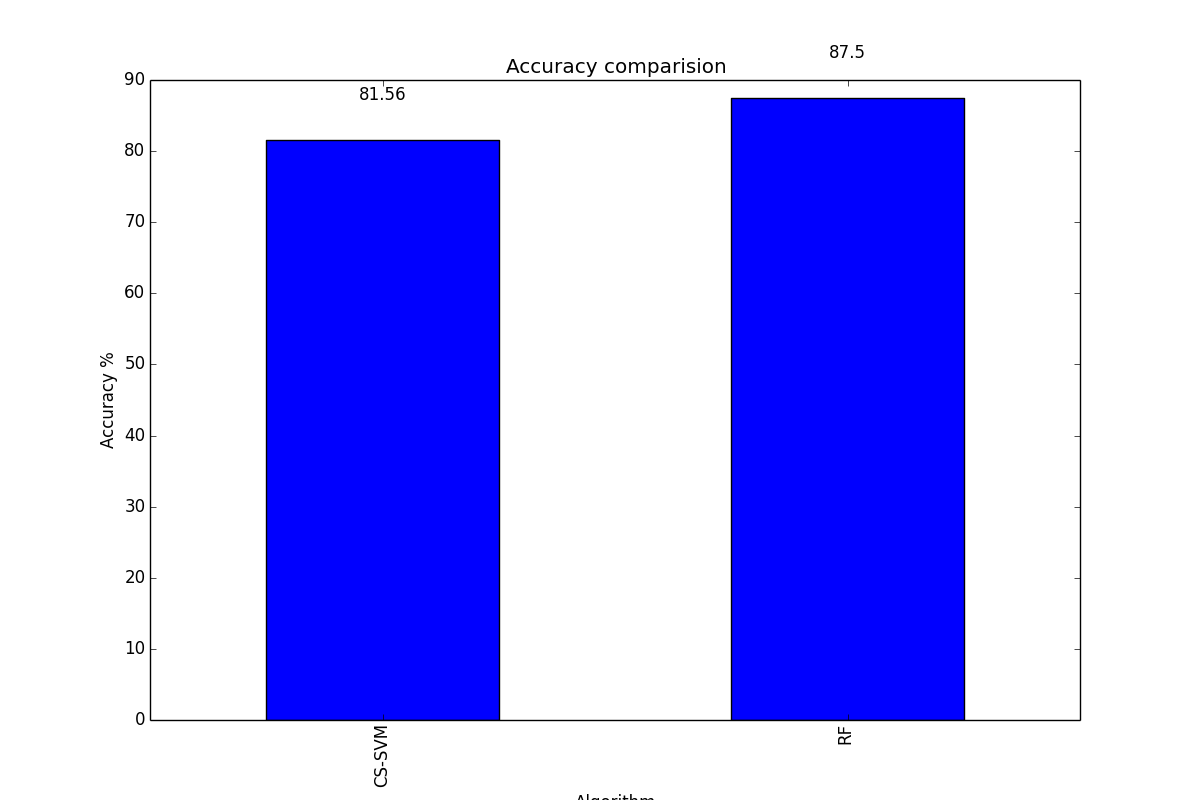}
    \mbox{Fig 17: Comparing Accuracies of CS-SVM as obtained in Devi(2015) and 
    RF for BSE-SENSEX}
    \end{center}
    \end{figure}

    SVM with Cuckoo search optimization performs really well giving accuracy 
    results of above 80\%. However, the Random Forest classifier still 
    performs better than the model proposed in \cite{devi}.
    \par
    From the comparative analysis we have done in this section, we can
    confidently say that our model outperforms the models as seen in 
    various papers in our literature survey. We believe that this is due 
    to the lack of proper data processing in 
    \cite{stanford1},\cite{stanford2},\cite{stanford3},\cite{devi}. In this 
    paper, we have performed exponential smoothing which is a rule of thumb
    technique for smoothing time series data. Exponential smoothing 
    removes random variation in the data and makes the learning process
    more easier. But in none of the papers we have reviewed, have used
    exponential smoothing to smooth their data. Another important
    reason could be the inherent non linearity in data. This fact
    discourages the use of linear classifiers. However in \cite{stanford1},
    the authors have used linear classifier algorithm: Logistic Regression
    as their supervised learning algorithm which yielded a success rate of
    55.65\%. We believe that the use of SVM in \cite{stanford2} and 
    \cite{stanford3} is not very wise. Due to that fact that the two classes
    in consideration (rise or fall) are linearly inseparable, researchers
    are compelled to use SVM with non linear kernels such as Gaussian kernel
    or Radial Basis Function. Despite many advantages of SVMs, from a 
    practical point of view, they have some drawbacks. An important 
    practical question that is not entirely solved, is the selection of the 
    kernel function parameters - for Gaussian kernels the width parameter 
    $\sigma$ - and the value of $\epsilon$ in the $\epsilon$ loss insensitive
    function (Horváth (2003) in Suykens et al.).

    \subsection{Conclusion}
    Predicting stock market due to its non linear, dynamic and complex 
    nature is really difficult. However in the recent years, machine 
    learning techniques have proved effective in stock forecasting. Many
    algorithms such as SVM, ANN etc. have been studied for robustness in
    predicting stock market. However, ensemble learning  methods have 
    remained unexploited in this field. In this paper, we have used 
    random forest classifier to build our predictive model and our model
    has produced really impressive results. The model is proved to be 
    really robust in predicting future direction of stock movement. The 
    robustness of our model has been evaluated by calculating various
    parameters such as accuracy, precision, recall and specificity. For
    all the datasets we have used i.e, AAPL, MSFT and Samsung, we were
    able to achieve accuracy in the range 85-95\% for long term 
    prediction. ROC curves were also plotted to evaluate our model. The 
    curves graphically proved the robustness of our model. It was also 
    proved that our classification algorithm converges as more number of 
    trees are added to the random 
    forest.
    \par
    Our model can be used for devising new strategies for trading or 
    to perform stock portfolio management, changing stocks according to
    trends prediction. For future work, we could build random forest 
    models to predict trends for short time window in terms of hours or 
    minutes. Ensembles of different machine learning algorithms can also
    be checked for its robustness in stock prediction. We also recommend 
    exploration of the application of Deep Learning practices in Stock 
    Forecasting. These practices involve learning weight coefficients on 
    large directed and layered graph. Deep Learning models, known earlier as 
    problematic in training, are now being embraced in stock price estimation due to the recent advances. 

    The model proposed indicates, for the first time, to the best of our 
    knowledge the nonlinear nature of the problem and the futility of 
    using linear discriminant type machine learning algorithms. The 
    accuracy reported is not pure chance but is based solidly on the 
    understanding that the problem is not linearly separable and hence 
    the entire suite of SVM type classifiers or related machine learning 
    algorithms should not work very well. The solution approach adopted 
    is a paradigm shift in this class of problems and minor modifications 
    may work very well for slight variations in the problem statement.


\begin{thebibliography}{20}
        \bibitem[Hellstrom and Holmstromm (1998)]{k} 
    Hellstrom, T., Holmstromm, K. (1998). Predictable Patterns in Stock Returns. 
    Technical Report Series IMa-TOM- 1997-09
    \bibitem[Saha, Routh and Goswami (2014)]{diffusion}
    Saha, Snehanshu., Routh, Swati., Goswami, Bidisha.(2014). Modeling Vanilla Option 
    prices: A simulation study by an implicit method. Journal of advances in 
    Mathematics, 6(1), 834-848.
    \bibitem[Widom (1995)]{asd}
    Widom, J. (1995). Research problems in data warehousing. In Proceedings of the 
    fourth international conference on information and knowledge management, CIKM '95 (
    pp. 25- 30). New York, NY, USA: ACM. 10.1145/221270.221319.
    \bibitem[Gencay (1999)]{stat1}
    R. Gencay, "Linear, non-linear and essential foreign exchange rate 
    prediction with simple technical trading rules," Journal of International 
    Economics, vol. 47,no.!, pp. 91-107,1999.
    \bibitem[Timmermann and Granger (2004)]{stat2}
    A. Timmermann and C. W Granger, "Efficient market hypothesis and 
    forecasting," Interational Journal of  Forecasting, vol. 20,no.!, pp. 15-
    27,2004.
    \bibitem[Bao and Yang (2008)]{stat3}
    D. Bao and Z. Yang, "Intelligent stock trading system by turning point 
    confirming and probabilistic reasoning," Expert Systems with Applications, 
    vol.34,no. 1,pp. 620-627,2008.
    \bibitem[Li, Li and Yang (2014)]{stanford1}
    Haoming Li, Zhijun Yang and Tianlun Li (2014). Algorithmic 
    Trading Strategy Based On Massive Data Mining. Stanford 
    University.
    \bibitem[Dai and Zhang(2013)]{stanford2}
    Yuqing Dai, Yuning Zhang (2013). Machine Learning in Stock Price Trend 
    Forecasting. Stanford University.
    \bibitem[Xinjie (2014)]{stanford3}
    Xinjie (2014). Stock Trend Prediction With Technical Indicators 
    using SVM. Stanford University.
    \bibitem[Geurts and Louppe (2011)]{randomizedtree}
    Pierre Geurts, Gilles Louppe . Learning to rank with extremely 
    randomized tree. JMLR: Workshop and Conference Proceedings 14 
    (2011) 49–61 
    \bibitem[Giacomel, Galante and Pareira (2015)]{ANN}
    Felipe Giacomel, Renata Galante, Adriano Pareira. An Algorithmic Trading 
    Agent based on a Neural Network Ensemble: a Case of Study in North 
    American and Brazilian Stock Markets. 2015 IEEE/WIC/ACM International 
    Conference on Web Intelligence and Intelligent Agent Technology
    \bibitem[Boonpeng and Jeatrakul (2016)]{dss}
    Sabaithip Boonpeng, Piyasak Jeatrakul (2016). Decision Support System for 
    Investing in Stock Market by using OAA-Neural Network. 8th International 
    Conference on Advanced Computational Intelligence Chiang Mai, Thailand; 
    February 14-16, 2016 
    \bibitem[Devi, Bhaskaran and Kumar (2015)]{devi}
    Ms. K. Nirmala Devi, Dr.V.Murali Bhaskaran, G. Prem Kumar (2015). Cuckoo 
    Optimized SVM for Stock Market Prediction.IEEE Sponsored 2nd International 
    Conference on Innovations in Information, Embedded and Communication 
    systems (ICJJECS)2015  
    \bibitem[Breiman (2001)]{rf}
    Leo Breiman (2001), Statistics Department
    ,University of California Berkeley, CA 94720. Random Forests. 
    \bibitem[Bylander and Hanzlik (1999)]{oob}
    Tom Bylander,Dennis Hanzlik (1999). Estimating Generalization Error 
    Using Out-of-Bag Estimates. AAAI-99 Proceedings. Copyright © 1999, AAAI 
    (www.aaai.org). All rights reserved.
    \bibitem[Fama and Malkiel(1970)]{EMH} 
    Malkiel, B. G. and Fama, E. F.(1970). 
    Efficient capital markets: A review of theory and empirical work.
    The Journal of Finance, 25(2), 383-417.
    \bibitem[Malkiel (2003)]{emhcrit} 
     Malkiel, Burton G. (2003). 
    The efficient market hypothesis and its critics.
    The Journal of Economic Perspectives, 17(1), 59-82.


        \end{thebibliography}
    \end{document}